\documentclass[letterpaper]{article} 
\usepackage{aaai2026}  
\usepackage{times}  
\usepackage{helvet}  
\usepackage{courier}  
\usepackage[hyphens]{url}  
\usepackage{graphicx} 
\usepackage{natbib}  
\usepackage{caption} 
\usepackage{algorithm}
\usepackage{algorithmic}

\usepackage{newfloat}
\usepackage{listings}
\DeclareCaptionStyle{ruled}{labelfont=normalfont,labelsep=colon,strut=off} 
\floatstyle{ruled}
\newfloat{listing}{tb}{lst}{}
\floatname{listing}{Listing}
\title{\Tool{}: Benchmarking LLMs on Specification Generation Tasks for Operating System Verification \thanks{This is an extended version of a paper published at AAAI-26.}}
\author{
    Shangyu Li\textsuperscript{\rm 1},
    Juyong Jiang\textsuperscript{\rm 1,\rm 2},
    Tiancheng Zhao\textsuperscript{\rm 3},
    Jiasi Shen\textsuperscript{\rm 1}
}
\affiliations{
    \textsuperscript{\rm 1}The Hong Kong University of Science and Technology\\
    \textsuperscript{\rm 2}The Hong Kong University of Science and Technology (Guangzhou)\\
    \textsuperscript{\rm 3}Georgia Institute of Technology\\
    \{sliew,jjiang472\}@connect.ust.hk, tzhao350@gatech.edu, sjs@cse.ust.hk
}

\usepackage{bibentry}

\usepackage{tcolorbox}
\usepackage{enumitem}
\usepackage{amsmath}
\usepackage{booktabs}
\usepackage{makecell}
\usepackage{multirow}
\usepackage{verbatim}
\usepackage{fancyvrb}
\usepackage{subcaption}
\DefineVerbatimEnvironment{pycode}{Verbatim}{%
formatcom=\fontfamily{zi4}\selectfont\small 
}

\definecolor{myblue1}{RGB}{218, 232, 252}
\definecolor{myblue2}{RGB}{108, 142, 191}
\definecolor{myblue3}{RGB}{153, 204, 255}

\tcbset{
    systembox/.style={
        colback=green!10,
        colframe=green!50,
        sharp corners,
        boxrule=1pt,
        fonttitle=\bfseries,
        coltitle=black,
        before skip=5pt,
        after skip=0pt,
    },
    assumptionbox/.style={
        colback=blue!10,
        colframe=blue!30,
        sharp corners,
        boxrule=1pt,
        fonttitle=\bfseries,
        coltitle=black,
        before skip=0pt,
        after skip=0pt,
    },
    programmingbox/.style={
        colback=orange!10,
        colframe=orange!50,
        sharp corners,
        boxrule=1pt,
        fonttitle=\bfseries,
        coltitle=black,
        before skip=0pt,
        after skip=0pt,
    },
    examplesbox/.style={
        colback=yellow!10,
        colframe=yellow!50!black,
        sharp corners,
        boxrule=1pt,
        fonttitle=\bfseries,
        coltitle=black,
        before skip=0pt,
        after skip=0pt,
    },
    questionbox/.style={
        colback=myblue3!50,
        colframe=myblue3!70,
        sharp corners,
        boxrule=1pt,
        fonttitle=\bfseries,
        coltitle=black,
        before skip=5pt,
        after skip=5pt,
    },
    answerbox/.style={
        colback=myblue2!10,
        colframe=myblue2!50,
        sharp corners,
        boxrule=1pt,
        fonttitle=\bfseries,
        coltitle=black,
        before skip=5pt,
        after skip=5pt,
    }
}

\newcommand{\autoref}[1]{Sec.~\ref{#1}}
\newcommand{\autofigref}[1]{Fig.~\ref{#1}}
\newcommand{\autotabref}[1]{Table~\ref{#1}}

\newcommand{\Tool}{\textsc{OSVBench}}

\begin{document}

\maketitle
\begin{abstract}
We introduce \Tool{}, a new benchmark for evaluating Large Language Models (LLMs) on the task of generating complete formal specifications for verifying the functional correctness of operating system kernels.
This benchmark is built upon a real-world operating system kernel, Hyperkernel, and consists of 245 complex specification generation tasks in total, each of which is a long-context task of about 20k-30k tokens.
The benchmark formulates the specification generation task as a program synthesis problem confined to a domain for specifying states and transitions. This formulation is provided to LLMs through a programming model.
The LLMs must be able to understand the programming model and verification assumptions before delineating the correct search space for syntax and semantics and generating formal specifications.
Guided by the operating system's high-level functional description, the LLMs are asked to generate a specification that fully describes all correct states and transitions for a potentially buggy code implementation of the operating system.
Experimental results with 12 state-of-the-art LLMs indicate limited performance of existing LLMs on the specification generation task for operating system verification.
Significant disparities in their performance highlight differences in their ability to handle long-context code generation tasks.

\begin{links}
    \link{Code}{https://github.com/lishangyu-hkust/OSVBench}
\end{links}

\end{abstract}

\section{Introduction}

Large Language Models (LLMs) have shown great potential in software engineering tasks, such as code generation \cite{austin2021program,athiwaratkun2022multi,zan2023large,jiang2024survey}, code summarization \cite{ahmed2024automatic}, and bug repair \cite{jin2023inferfix}. 
An important aspect of software engineering is software verification.
Software verification uses rigorous mathematical reasoning to prove the absence of bugs in software \cite{dahl1972structured}, which is essential in ensuring the correctness of software in safety-critical domains such as aerospace, healthcare, and nuclear energy \cite{klein2009sel4, amani2016cogent, o2016refinement}, where software bugs could lead to catastrophic economic losses or even endanger human lives.

A class of software for which verification is especially valuable is that of operating system (OS) kernels, which are fundamental components of many critical infrastructures.
Yet the inherent complexity, concurrency, and hardware interactions of OS kernels render the verification process highly challenging.
Manual software verification requires advanced knowledge of formal methods and program analysis, and there is a shortage of professionals who are capable of conducting such verification \cite{klein2009sel4}.
Verifying the seL4 microkernel \cite{klein2009sel4} required 11 person-years of effort for 10k lines of C code, and verifying two operations of the BilbyFs file system required 9.25 person-months of effort \cite{amani2016cogent} for 1,350 lines of code.
These challenges highlight the need for automation.

Formally verifying an OS kernel often involves (1) defining formal specifications that characterize the properties that must be satisfied by the kernel \cite{klein2009sel4, chajed2022concurrent, chen2015fscq, chen2017crash} and (2) constructing formal proofs with a theorem prover to show that the code implementation satisfies the specifications.
Existing research focuses mainly on automating proof generation \cite{chen2024automated, zhang2024selene} and often overlooks the crucial process of specification generation \cite{sammler2021refinedc, leino2010dafny, jacobs2008verifast, ma2024specgen}.
Generating specifications is especially challenging for OS kernel verification because such specifications are often ad hoc \cite{chen2017crash, chen2015fscq, chajed2022concurrent, sigurbjarnarson2016pushfs} and require substantial domain expertise.
For instance, creating the formal specification for seL4 \cite{klein2009sel4} required 7 person-months of effort.

We introduce \Tool{}, a benchmark suite for evaluating the capabilities of LLMs in generating formal specifications for verifying the functional correctness of an OS kernel.
\Tool{} is derived from the Hyperkernel project \cite{hyperkernel} and consists of 245 specification generation tasks, each of which asks an LLM to generate a specification based on the code implementation of a system call in the OS kernel along with its natural language functional description.
To capture realistic scenarios where OS kernels may have vulnerabilities, we inject five types of bugs into the code implementation before providing it to the LLMs.

Each task in \Tool{} represents a complex and intricate program synthesis problem with a long context of approximately 20k to 30k tokens.
This design enables us to investigate the capabilities of LLMs in understanding and manipulating extensive contextual information and domain knowledge.

We evaluate 12 state-of-the-art LLMs on \Tool{}.
The experimental results show that the LLMs exhibit limited performance in automating formal specification generation.
Significant disparities in their performance on the benchmark highlight differences in their ability to handle long-context code generation tasks.
We also discuss the impact of varying types and quantities of injected bugs on the quality and effectiveness of the generated specifications.

\section{Related Work}

\paragraph{LLM for software verification.}
Software verification \cite{d2008survey} ensures that software conforms to specified properties or requirements, playing a critical role in guaranteeing software reliability and correctness.
In this domain, operating system kernel verification \cite{klein2014comprehensive} has been a central research goal for ensuring the reliability and security of critical software systems. Early foundational work includes efforts such as UCLA Secure Unix \cite{walker1980specification}, PSOS \cite{feiertag1977proving}, and KIT \cite{bevier1989kit}, which laid the groundwork for formal approaches to kernel correctness.
Recent progress has expanded to leverage formal methods like theorem proving \cite{hyperkernel} and model checking \cite{klein2009sel4}, aiming for high-assurance kernels with mathematically verified properties.
Moreover, prior work on leveraging LLMs for software verification mainly focused on generating proofs from specifications \cite{chen2024automated, zhang2024selene}, which involves translating one formal semantic representation (specifications, in various forms) into another (proofs expressed in formal languages).
Some studies focus on automatically generating verification code from given code snippets and their pre- and postconditions to prove that they satisfy a specified set of properties~\cite{loughridge2024dafnybench, 10.1145/3643763}, whereas others design a consistency-checking loop among code, docstrings, and formal annotations to filter out incorrect code~\cite{sun2024clover}.
Although some studies have explored the task of specification generation, much of this work has focused on general-purpose specification generation \cite{ma2024specgen, cao-etal-2025-informal}, which differs significantly from the generation of OS kernel specifications due to the distinct verification assumptions and requirements encountered in this domain.

\paragraph{LLM for code generation.}
Formal specifications are a special form of source code.
Code generation and program synthesis have been extensively studied over the past few decades \cite{alur2013syntax, solar2008program, solar2006combinatorial, konure, zhang2024siro, sprout}.
The use of LLMs for code generation has gained significant attention in recent years, as these models have demonstrated remarkable capabilities in synthesizing code snippets from natural language descriptions \cite{austin2021program,athiwaratkun2022multi,zan2023large,jiang2024survey}.
Several studies have explored the potential of models in various coding tasks, ranging from simple function generation \cite{chen2021evaluating,luo2023wizardcoder} to more complex programming challenges~\cite{jimenez2023swe,ding2024crosscodeeval,li2024infibench, khatry2025crustbenchcomprehensivebenchmarkctosaferust, complexcodeeval, tan2025profix, wei-etal-2025-satbench, wei-etal-2025-equibench}.
Despite these advances, code generation for specific domains, such as OS kernel verification, poses unique challenges that are not fully addressed by existing general-purpose LLMs. The complexity and uniqueness of the syntax and semantics involved require models not only to understand programming languages but also to grasp domain-specific knowledge and verification assumptions.
This challenge calls for a benchmark that evaluates the capabilities of LLMs in generating specifications for OS kernel verification.

\paragraph{LLM for static and dynamic program analysis.}
Existing LLM-based static analysis techniques often rely on prompting LLMs to perform source-sink reachability analyses on programs \cite{wang2024llmdfa, wang2024sanitizing, wang2024llmsa, li2024llm}. However, limited research has explored LLM-based static analysis specifically for the Linux kernel, which presents a significant challenge due to its long-context reasoning requirements. This complexity arises from the kernel's intricate call graphs and alias relationships. In contrast, dynamic analysis techniques, such as fuzzing, have seen broader application of LLMs across various domains, including smart contracts \cite{shou2024llm4fuzz}, the Linux kernel~\cite{yang2023kernelgpt}, and universal domains~\cite{xia2024fuzz4all}.

\begin{figure*}[t]
    \centering
    \includegraphics[width=\linewidth]{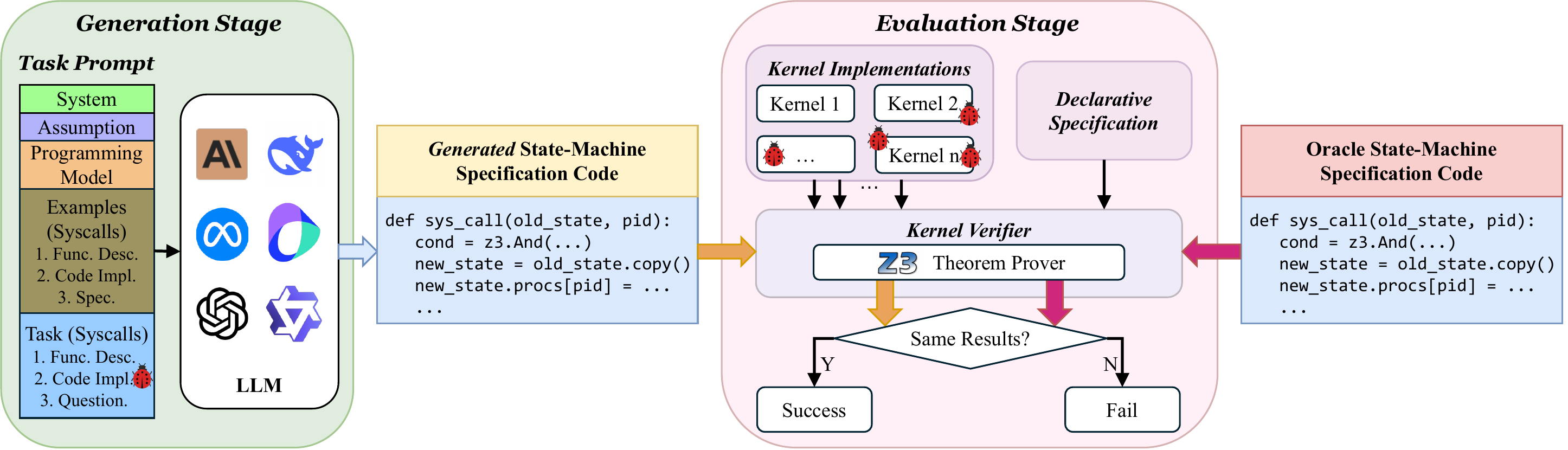}
    \caption{The workflow of the \Tool{} benchmark suite.
The workflow consists of the generation stage and the specification quality evaluation stage.
During the generation stage, the input to LLMs includes a system prompt, verification assumptions, a programming model, system call (syscall) examples, and a task question (see \autoref{sec:benchmark-task-construction} for details).
Based on this input, the LLMs are tasked with generating the correct state-machine specification for a given syscall.
The evaluation stage checks the correctness of the generated state-machine specification by comparing the verification results it produces against those produced by a ground-truth oracle state-machine specification when both are fed into a kernel verifier. In particular, the kernel verifier takes as input a state-machine specification and a formal declarative specification of the kernel (see \autoref{sec:hyperkernel-preliminaries}), then uses them to perform verification (see \autoref{sec:problem-formulation}) on a potentially buggy kernel implementation (see \autoref{sec:benchmark-task-construction}). If both state-machine specifications produce the same verification results for all kernel implementations, the generated one is considered correct. If the two results differ for any kernel implementation, the generated state-machine specification is considered incorrect.
}
    \label{fig:workflow}
\end{figure*}

\section{\Tool{}}

\autofigref{fig:workflow} presents the \Tool{} workflow.
We introduce key concepts and outline how we construct the benchmark tasks.

\subsection{Preliminaries}
\label{sec:hyperkernel-preliminaries}

\paragraph{Hyperkernel and its verifier.}
Hyperkernel \cite{hyperkernel} is an OS kernel verification project that includes both a real-world kernel implementation and a verification framework built on the automated theorem prover Z3 \cite{de2008z3}. The kernel implementation consists of 50 system calls, among which 49 are supported by its verifier. These syscalls cover key functionalities such as process management, virtual memory, file descriptors, device interaction, inter-process communication, and scheduling. The entire codebase, including both the kernel implementation and associated user-space components, consists of approximately 18,000 lines of C and assembly code.

Hyperkernel is particularly suitable as a challenging research vehicle for evaluating specification generation tasks for the following reasons:
(1) Hyperkernel adopts a standardized approach \cite{klein2010refinement} to modeling kernel execution as a state machine.
(2) Crafting these specifications requires significant expertise and non-trivial effort.
(3) Hyperkernel employs an automated theorem prover, Z3, instead of an interactive theorem prover \cite{isabelle,coq,dafny} to formally verify the functional correctness of the OS kernel.
The automated theorem prover streamlines the verification process for performing specification generation tasks.

\paragraph{Two types of specifications.}
To verify a kernel implementation, the verifier for Hyperkernel requires two types of specifications as input to ensure functional correctness. The first is a state-machine specification that defines the intended behavior, especially state transition behavior, of the OS kernel. The second is a higher-level declarative specification that outlines the overarching properties and invariants that any state-machine specification must satisfy.
For example, one such invariant is that a page with an owner is not free, denoted as:
\begin{multline}
\label{equation:page}
\forall \sigma \in \Sigma_{\mathit{Spec}}, \mathit{pn} \in \sigma, \mathsf{is\_valid}(\mathit{pn}, \sigma) \Rightarrow \\
(\mathsf{is\_valid}(\sigma_{\mathsf{page\_owner}}[\mathit{pn}], \sigma) \Leftrightarrow (\sigma_{\mathsf{page\_type}}[\mathit{pn}] \not = \tau_{\mathit{free}})),
\end{multline}
where $\sigma$ denotes a specific kernel state defined in the state-machine specification, $\mathit{pn}$ denotes a page number under the current kernel state $\sigma$, $\sigma_\mathsf{page\_owner}$ and $\sigma_\mathsf{page\_type}$ are mappings from a page to its owning process and its type, respectively, and $\tau_{\mathit{free}}$ represents the type of the freed page. The $\mathsf{is\_valid}$ function checks the validity of a given process.

The verifier establishes two theorems. The first one proves that the kernel implementation is a refinement of the state-machine specification.
The refinement relation states that a state transition in the kernel implementation is equivalent to a state transition in the specification.
The second theorem demonstrates that all reachable states in the state-machine specification must satisfy the properties and invariants defined in the declarative specifications.
The theorem is expressed as:
\begin{equation}
\label{equation:declarative-spec}
\forall \sigma_\mathit{spec}, \mathit{i}, P(\sigma_\mathit{spec}) \Rightarrow P(t_\mathit{spec}(\sigma_\mathit{spec}, \mathit{i})),
\end{equation}
where $\sigma_\mathit{spec}$ denotes a kernel state in the state-machine specification, $\mathit{i}$ represents the input for a state transition $t_\mathit{spec}$, and $P$ is the predicate or invariant defined in the declarative specification \cite{hyperkernel}.

\subsection{Problem Formulation}
\label{sec:problem-formulation}
In the specification generation task, the LLM takes a prompt specifically designed for a syscall and synthesizes a state-machine specification for the syscall.
Once the LLM synthesizes the state-machine specification, we feed it into the Hyperkernel verifier. The verifier has access to the declarative specification and a potentially buggy implementation for the same syscall in the kernel. The verifier checks the equivalence between the state-machine specification and the syscall in the kernel implementation by performing symbolic execution \cite{cadar2008klee} on the compiled LLVM IR \cite{lattner2004llvm} of the kernel implementation and invoking the Z3 theorem prover.
An equivalent result indicates that the state-machine specification accurately characterizes the functionality of the syscall implementation.
An inequivalent result indicates an inconsistency between the two; that is, there may be a bug in the implementation or the state-machine specification may not accurately characterize the syscall functionality.
In either case, if the result for an LLM-generated state-machine specification differs from the result for a ground-truth oracle state-machine specification, the generated one is considered incorrect.

\paragraph{Formulating as a general-purpose program synthesis problem.}

It is challenging for LLMs to directly solve the specification generation task due to the inherent complications in formally verifying OS kernels.
We therefore reformulate the specification generation task as a general-purpose program synthesis problem:
\begin{itemize}
    \item We introduce explicit verification assumptions by systematically documenting common assumptions related to the OS kernel, such as hardware behavior and memory layout. This process is important because many such assumptions have been used implicitly by OS kernel verifiers such as Hyperkernel, which makes it infeasible for LLMs to directly perform reliable verification.

    \item We introduce fixed declarative specifications to enable LLMs to focus solely on generating state-machine specifications. This restriction is important because the declarative specification enforces the correctness of the state-machine specification. If we had instead tasked LLMs with generating both the state-machine specification and the declarative specification, as with the original Hyperkernel design, it would have been more difficult to automatically determine whether the generated specifications are correct---in this case, a successful verification result could have been due to an incorrect declarative specification that fails to identify errors.

    \item We introduce a deterministic synthesis domain with known constants extracted from the compiled LLVM IR of the kernel implementation, utility functions, Z3 functions, and Python classes to restrict the search space of syntactically correct specifications.
Defining this domain is important because the state-machine specifications often use such constants and external functions that are not visible in kernel code implementations.
By providing them as known building blocks and making them apparent, we make it more tractable for LLMs to generate a correct specification.

\end{itemize}

\begin{figure}
\footnotesize
    \centering
\begin{tcolorbox}[    colback=gray!10,
    colframe=white,
    boxrule=0mm,
    width=\linewidth,
    sharp corners,
    colbacktitle=white,
    boxsep=0mm,
    left=1mm,
    right=1mm,
    top=0mm,
    bottom=3mm
    ]
\begin{align*}
&\mathsf{Specification} := \mathsf{State} \\ 
&\mathsf{State} := \mathsf{Param} \mid \mathit{if}~ \mathsf{Cond}~ \mathsf{State} \ 
 \mathsf{State} \mid \\ 
& \quad \quad \quad \quad \mathsf{State} .(\mathit{field_i} \leftarrow \mathsf{Expression})^{+} \\
&\mathsf{Expression} := \mathsf{Value} \mid \mathsf{Expression} \ \mathit{aop} \ \mathsf{Expression} \mid {} \\
&\quad \quad \quad \quad \mathit{if} \ \mathsf{Cond} \ \mathsf{Expression} \  \mathsf{Expression} \\
&\mathsf{Value} := \mathsf{Param} \mid \mathsf{Const} \mid \mathsf{State}.\mathit{field_i} \\ 
&\mathsf{Cond} := \mathsf{Value} ~ \mathit{lop} ~ \mathsf{Value} \mid \mathsf{Cond} \wedge \mathsf{Cond} \mid \ \mathsf{Cond} \vee \mathsf{Cond} \\
&\mathit{aop} := \texttt{+} \mid \texttt{-} \mid \times \mid \div  \\
&\mathit{lop} := \texttt{==} \mid \texttt{!=} \mid \texttt{>} \mid \texttt{<} \mid \texttt{>=} \mid \texttt{<=}   
\end{align*}
\end{tcolorbox}
\caption{Abstract syntax for state-machine specifications.
A specification is a kernel state. A kernel state may be an input parameter of the system call, a conditional state that branches to one of two existing states according to a condition, or a modified state that assigns new expressions to the attributes of an existing state.
An expression may be an existing value, an arithmetic expression, or a conditional expression.
A value may be an input parameter, a predefined constant literal, or an attribute of an existing state.
A condition may be a logical expression, a conjunction, or a disjunction.
Intuitively, a specification defines how to transition the OS kernel to a subsequent state.
The specification defines state transition and field assignments, along with the conditions that must be satisfied.
}
    \label{fig:spec:domain}
\end{figure}

\begin{table*}[t]
    \centering
    \caption{Performance comparison (\textit{Pass@1} \%) of various models with a 5-shot prompt. \textit{Pass@1} evaluates performance in the single-attempt setting: the model generates one specification, and success is recorded only if that specification is correct. Models marked with $^*$ are reasoning LLMs, while ${\triangle}$ denotes closed-source models; all models without ${\triangle}$ are open-source.
    The columns from \textbf{Incorrect Pointer} to \textbf{Bounds Checking} correspond to specific types of bugs injected into the syscall code within the task prompts. The column \textbf{Correct} corresponds to cases where the provided code implementations are bug-free. Lastly, the column \textbf{Total} presents the overall \textit{Pass@1} rate across all 245 tasks.
}
    \resizebox{0.9\linewidth}{!}{
    \begin{tabular}{ll|cccccc|c}
        \toprule
        \textbf{Institution} & \textbf{Model} & \makecell[c]{\textbf{Incorrect}\\ \textbf{Pointer}} &  \makecell[c]{\textbf{Incorrect}\\ \textbf{Privilege}} & \makecell[c]{\textbf{Memory}\\ \textbf{Leak}} & \makecell[c]{\textbf{Buffer}\\ \textbf{Overflow}} & \makecell[c]{\textbf{Bounds}\\ \textbf{Checking}} & \textbf{Correct} & \textbf{Total} \\ 
        \midrule
        \multirow{3}{*}{OpenAI} 
        & o1$^{*\triangle}$ & 12.68    & 21.43 & 13.51 & 20.37 & 23.15  & 28.57 & 23.67 \\ 
        & o3-mini$^{*\triangle}$ & 19.72 & 18.75 & 18.92 & 12.96
        & 15.74 & 26.53 & 22.04 \\ 
        & GPT-4o$^{\triangle}$ & 33.80 & 34.82 & 32.43 & 33.33 & 36.11 & 42.86 & 38.78 \\ 
        \midrule
        \multirow{2}{*}{DeepSeek} 
        & DeepSeek-R1$^*$ & 32.39 & 21.43 & 13.51 & 20.37 & 23.15 & 42.86 & 40.82 \\
        & DeepSeek-Chat & 38.02 & 39.29 & 36.49 & 44.44 & 43.52 & 51.02 &  46.53 \\ 
        \midrule
        \multirow{2}{*}{Meta} & Llama-3.1-70B-Instruct & 12.68 & 18.75 & 12.16 & 16.67 & 22.22 & 22.45 &  22.45 \\ 
        & Llama-3.1-8B-Instruct & 0.00 & 11.61 & 0.00 & 12.96 & 9.26 & 10.20 &  10.61 \\ 
        \midrule
        \multirow{3}{*}{Qwen Team} & QwQ-32B-Preview$^*$ & 14.08 & 23.21 & 20.27 & 20.37 & 23.15 & 22.45  & 24.08 \\
        & Qwen2.5-72B-Instruct & 25.35 & 26.79 & 24.32 & 25.93 & 30.56 & 34.69 &  32.24 \\ 
        & Qwen2.5-Coder-7B-Instruct & 0.00 & 8.04 & 0.00 & 3.70 & 5.56 & 4.08 & 4.90 \\ 
        \midrule
        Anthropic & Claude-3.5-sonnet$^{\triangle}$ & 39.44 & 41.96 & 39.19 & 48.15 & 39.81 & 46.94 & 44.90 \\ 
        \midrule
        ByteDance & Doubao-1.5-pro$^{\triangle}$ & 50.70 & 48.21 & 45.95 & 40.74 & 52.78 & 63.27 & 55.10 \\ 
        \bottomrule
    \end{tabular}
    }
\label{table:main-results}
\end{table*}

\begin{table*}[t]
    \centering
    \medskip
    \caption{
    Comparison of two-round self-repair performance (repair success rate \%) for models using a 5-shot prompt.
    The columns \textbf{Syntax Error} and \textbf{Semantic Error} present the repair success rates for specifications that produce the types of errors. The columns from \textbf{Incorrect Pointer} to \textbf{Correct} present the repair success rates for specifications derived from the task prompt, categorized by the types of buggy code implementations in the prompt.
    The \textbf{Total} column presents the overall repair success rate for all erroneous specifications addressed in each round. Values outside parentheses are the success rates in the first repair round, while values within parentheses are the success rates in the second round.
    }
    \resizebox{\linewidth}{!}{
    \begin{tabular}{l|cccccccc|c}
        \toprule
        \textbf{Model} & \makecell{\textbf{Syntax}\\ \textbf{Error}} & \makecell{\textbf{Semantic}\\ \textbf{Error}} & \makecell[c]{\textbf{Incorrect}\\ \textbf{Pointer}} &  \makecell[c]{\textbf{Incorrect}\\ \textbf{Privilege}} & \makecell[c]{\textbf{Memory}\\ \textbf{Leak}} & \makecell[c]{\textbf{Buffer}\\ \textbf{Overflow}} & \makecell[c]{\textbf{Bounds}\\ \textbf{Checking}} & \textbf{Correct} & \textbf{Total} \\ 
        \midrule
        GPT-4o & 16.47 (1.25) & 12.31 (2.08) & 17.02 (5.13) & 17.81 (3.33) & 6.12 (2.17) & 8.33 (3.03) & 17.39 (1.75) & 17.86 (0.00) & 14.67 (1.56) \\ 
        DeepSeek-Chat & 8.57 (6.25) & 14.75 (0.00) & 2.27 (4.65) & 8.82 (1.61) & 6.52 (0.00) & 10.00 (0.00) & 6.56 (3.51) & 20.83 (0.00) & 11.45 (2.59) \\ 
        
        \bottomrule
    \end{tabular}
    \medskip
    }
\label{table:self-repair}
\end{table*}

\autofigref{fig:spec:domain} defines the abstract syntax for the state-machine specifications that LLMs need to synthesize.
We define the specification generation task as a general-purpose program synthesis task, which takes a prompt as input and produces a synthesized state-machine specification code as output.
The synthesized specification (1) must conform to the specified programming model and (2) is used to verify the functional correctness of an OS kernel.

This synthesis problem is programming-language-agnostic and is particularly challenging in the following aspects:
(1) accurately mapping the semantics of functional descriptions to the corresponding regions in the specification,
(2) exhaustively considering all possible kernel states when there are a potentially divergent number of conditions, and (3) processing contextual information that averages 20k to 30k tokens, which requires advanced long-context learning capabilities.
We present concrete examples of these challenges in the extended version in Appendices~A and~E.

\subsection{Benchmark Task Construction}
\label{sec:benchmark-task-construction}

\paragraph{Prompt design.}
As illustrated in \autofigref{fig:workflow}, the prompt for each task is specifically designed for a particular syscall in the OS kernel.
Example task prompts are available in
Appendix~D of the extended version.
The prompt is structured into five components:
\begin{itemize}
    \item 
The system prompt describes the layout of contents and sections in the prompt and the general task to be solved.
\item 
The verification assumptions define the assumptions required to formally verify an OS kernel, such as the kernel running on a uniprocessor system and not providing multicore support.
\item
The programming model follows the standardized approach \cite{klein2010refinement} to modeling kernel execution as a state machine, using a set of Python classes and constants extracted from the compiled LLVM IR of Hyperkernel.
\item 
Few-shot syscall examples are provided, each of which consists of a functional description, a code implementation, and a corresponding oracle state-machine specification. These examples are carefully selected to ensure their representativeness.
\item
The task question consists of a syscall's functional description, a potentially buggy code implementation, and a question that asks the LLM to generate the corresponding state-machine specification.
\end{itemize}

\paragraph{Constructing functional descriptions.}

Note that both the few-shot syscall examples and the task question contain a functional description of the syscall.
For each of the 49 system calls in Hyperkernel, we manually drafted multiple versions of its functional description based on potential understandings of its intended behavior, purpose, and interactions within the OS kernel.
Then, we selected the best version based on clarity and technical precision \cite{hao2023syzdescribe}.

\paragraph{Systematic bug injection.}
We started with a correct implementation of the OS kernel and systematically generated a set of buggy implementations to reflect real-world scenarios where OS kernel implementations are not guaranteed to be correct and may contain various types and numbers of bugs.
In particular, we randomly injected five types of real-world bugs that were previously found in the xv6 kernel~\cite{cox2011xv6}:

\begin{itemize}
    \item Incorrect pointer: This bug means the kernel uses an unintended pointer value to access or update critical structures. It will cause incorrect state transitions, context corruption, and potential crashes.
    \item Incorrect privilege: This bug means privilege separation is broken, allowing user-space to perform privileged operations (e.g., I/O) directly. It will cause privilege escalation, device misuse, and bypass of kernel controls.
    \item Memory leak: This bug means allocated memory becomes unreachable and is not freed. It will cause cumulative resource loss, fragmentation, and eventual exhaustion.
    \item Buffer overflow: This bug means reads or writes exceed the bounds of a buffer or array. It will cause out-of-bounds memory access, corruption of kernel data, instability, and security vulnerabilities.
    \item Bounds checking: This bug means input validation fails to prevent invalid indices or sizes in corner cases. It will cause bypass of checks, out-of-bounds access, and undefined behavior.
\end{itemize}
\noindent These bugs are also prevalent in other OS kernels, such as the Linux kernel \cite{linux_cve_announce}.
We present example bugs in Appendix~B of the extended version.

To evaluate the impact of various bugs on the performance of LLMs in generating accurate state-machine specifications, we injected bugs into task questions. 
Specifically, we created a total of 245 specification generation tasks, each using a correct high-level functional description of a syscall paired with its potentially buggy code implementation. Among the 245 code implementations, 49 are correct, while the others contain 1--5 injected bugs.
We chose this small number of injected bugs because the number of severe vulnerabilities in mature OS kernels is often relatively small.

\section{Evaluation}

We use \Tool{} to evaluate the performance of a range of state-of-the-art large language models (LLMs) in the task of generating OS kernel verification specifications.

\begin{figure*}[t]
\centering
\includegraphics[width=\linewidth]{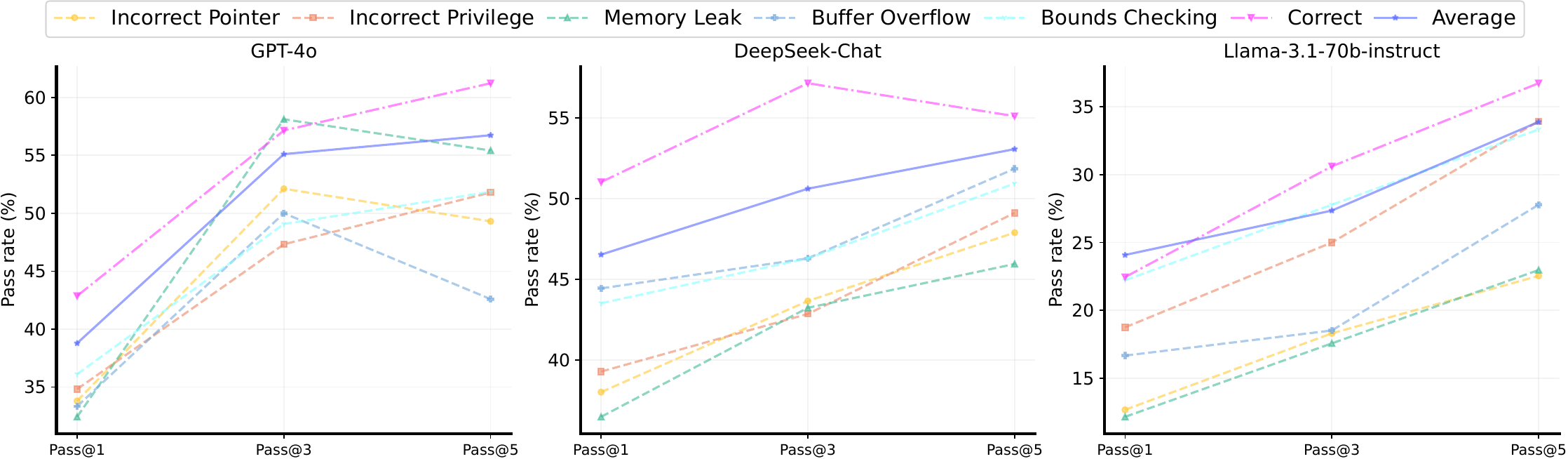}
\caption{Performance comparison of \textit{Pass@1}, \textit{Pass@3}, and \textit{Pass@5} of various models. The average performance of the models exhibits consistent improvement as k increases in \textit{Pass@k}.}
\label{fig:passk-results}
\end{figure*}
\begin{figure*}[t]
\centering
\includegraphics[width=\linewidth]{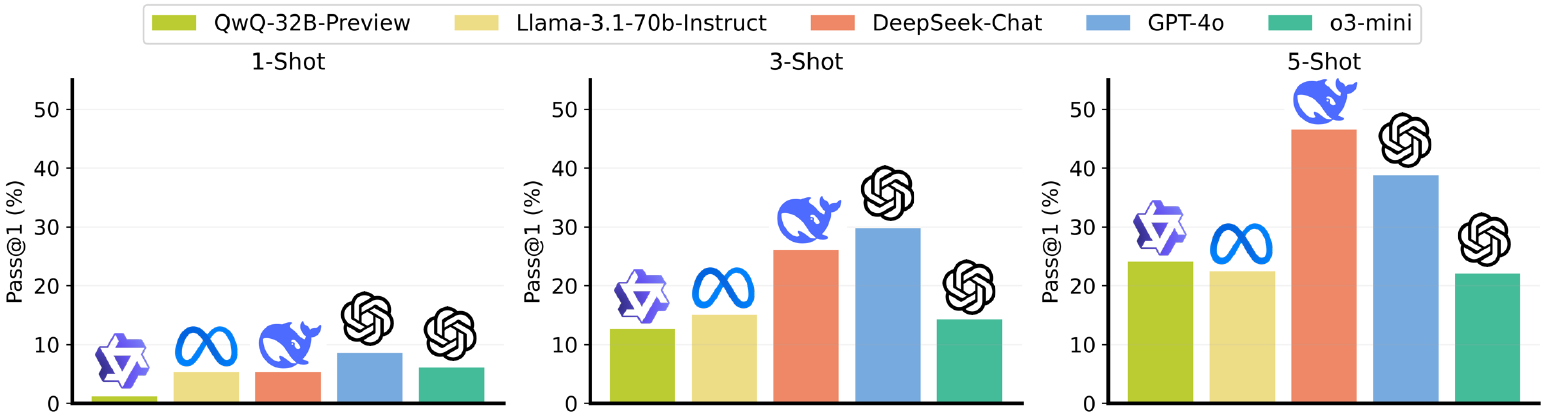}
\caption{Performance of various models with different numbers of few-shot examples.
}
\label{fig:few_shot_num}
\end{figure*}

\subsection{Experimental Setup}
\label{experiment-setup}
\paragraph{State-of-the-art LLMs.}
We evaluate existing LLMs developed by six leading institutions: OpenAI, DeepSeek, Meta, Anthropic, ByteDance, and the Qwen Team. Specifically, our evaluation includes the o1, o3-mini, and GPT-4o models from OpenAI; the DeepSeek-R1 and DeepSeek-Chat models from DeepSeek; the Llama-3.1-70B-Instruct and Llama-3.1-8B-Instruct models from Meta; the QwQ-32B-Preview, Qwen2.5-72B-Instruct, and Qwen2.5-Coder-7B-Instruct models from the Qwen Team; Claude-3.5-sonnet from Anthropic; and Doubao-1.5-pro from ByteDance.
These LLMs vary in key characteristics such as the number of parameters, open-source availability, data cutoff dates, and pretraining objectives. For all models, we employ a greedy search decoding strategy with \textit{Pass@1} to ensure the consistency of the evaluation.
To ensure a fair comparison, we use official provider APIs to access the models, measuring the \textit{Pass@1} for each model three times and reporting the average results.

\paragraph{Specification quality metrics.}
\label{paragraph:metrics}
To quantify the performance of LLMs on the tasks, we adopt the following metrics.
The metric \textit{Pass@k} indicates that at least one of the k generated specifications is correct, meaning the inconsistencies it identifies in all OS kernel implementations align with those detected by the oracle specification.
A \textit{Syntax error} occurs when the generated specification fails to execute correctly or terminates with an exception.
A \textit{Semantic error} occurs when there are no syntax errors, i.e., the verifier successfully translates the specification into an SMT formula \cite{de2008z3} and performs verification on the OS kernel implementation, but the pinpointed inconsistencies differ from those specified by the oracle specification.

\begin{table}[t]
    \caption{Syntax and semantic error rates (\%) in the specifications generated by LLMs across all 245 tasks. $^*$ denotes reasoning LLMs.
    A lower error rate indicates better performance.
    ${\triangle}$ denotes closed-source models, while unmarked models are open-source.
    }
    \resizebox{\linewidth}{!}{
    \begin{tabular}{l|cc}
        \toprule
        \textbf{Model} & \makecell[c]{\textbf{Syntax Error}} &  \makecell[c]{\textbf{Semantic Error}} \\ 
        \midrule
        o1$^{*\triangle}$      & 52.65     & 23.67 \\ 
        o3-mini$^{*\triangle}$ & 51.02     & 26.94 \\  
        GPT-4o$^{\triangle}$      & 35.10     & 26.53 \\ 
        \midrule
        DeepSeek-R1$^{*}$     & 32.65     & 26.53 \\ 
        DeepSeek-Chat       & 31.02    & 24.90 \\  
        \midrule
        Llama-3.1-70B-Instruct  & 44.90     & 32.65 \\  
        Llama-3.1-8B-Instruct   & 67.76     & 23.67 \\ 
        \midrule
        QwQ-32B-Preview$^*$     & 66.53     & 9.39 \\ 
        Qwen2.5-72B-Instruct    & 42.25     & 25.31 \\ 
        Qwen2.5-Coder-7B-Instruct   & 86.12     & 11.02 \\ 
        \midrule
        Claude-3.5-sonnet$^{\triangle}$    & 22.45     & 32.65 \\ 
        \midrule
        Doubao-1.5-pro$^{\triangle}$    & 23.67     & 21.22 \\ 
        \bottomrule
    \end{tabular}
    }
\label{tab:syntax_semantic_error}
\end{table}

\subsection{Main Results}
\autotabref{table:main-results} presents results on the performance of LLMs across institutions and bug categories.
The best-performing closed-source LLM, Doubao-1.5-pro, outperforms the best-performing open-source LLM, DeepSeek-Chat, with the highest average \textit{Pass@1} rate (55.10\%) and a superior ability to generate specifications for correct implementations (63.27\%), showcasing robust performance across all bug types.

In our experiments, models with more parameters outperform their smaller counterparts. For instance, Llama-3.1-8B-Instruct and Qwen2.5-Coder-7B-Instruct exhibit significantly weaker performance compared to their larger counterparts, Llama-3.1-70B-Instruct and Qwen2.5-72B-Instruct.
In 53 of 60 (88.3\%) experiments with injected bugs, the performance results are lower than those without bugs.
For example, memory leak bugs have the most pronounced effect on the DeepSeek-R1 model, while incorrect pointer bugs most significantly impact the o1 model.
These results indicate performance degradation caused by the presence of various types of bugs.

Contrary to common belief, widely regarded reasoning models, such as o1 and DeepSeek-R1, do not consistently outperform other models in this task. In particular, o1 demonstrates weak performance, performing worse than QwQ-32B-Preview, which challenges assumptions about the superiority of certain reasoning models in these tasks.
We speculate that the advanced reasoning models used in our experiments produce lengthy chains of reasoning traces, which could pose challenges to the long-context learning capabilities in OS verification scenarios.

\paragraph{Pass@k performance.}
We evaluate the performance of \textit{Pass@k} for GPT-4o, DeepSeek-Chat, and Llama-3.1-70B-Instruct.
\autofigref{fig:passk-results} presents the results.
For all three models, the average pass rates improve when k increases from 1 to 3 to 5.
Although \textit{Pass@1} for GPT-4o is lower than that for DeepSeek-Chat, GPT-4o outperforms DeepSeek-Chat at k = 3 and k = 5.

\subsection{Error Analysis and Self-Repair}

\autotabref{tab:syntax_semantic_error} presents the syntax and semantic error rates during the experiments in \autotabref{table:main-results}. The worst-performing LLM, Qwen2.5-Coder-7B-Instruct, is more prone to syntax errors than the best-performing LLM, Doubao-1.5-pro. This is evidenced by the higher semantic-to-syntax error rate ratio for Doubao-1.5-pro (21.22\% / 23.67\%) relative to Qwen2.5-Coder-7B-Instruct (11.02\% / 86.12\%).
\begin{figure}[t]
    \centering
    \includegraphics[width=\linewidth]{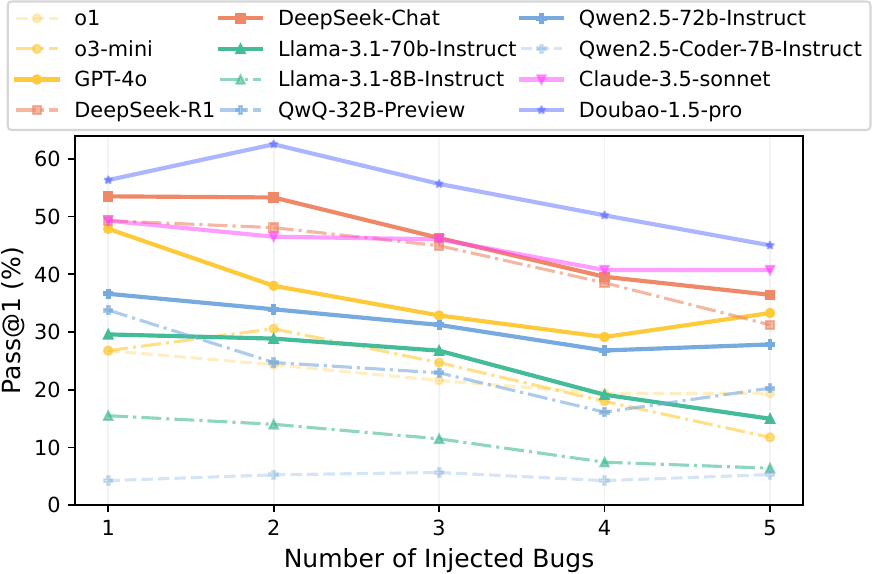}
    \caption{
    Performance comparison (\textit{Pass@1} \%) of 12 LLMs on specification generation tasks using syscall implementations injected with varying numbers of bugs.
    }
    \label{fig:bug_num}
\end{figure}

To further evaluate the error-handling capabilities of LLMs, we conduct a two-round self-repair process for GPT-4o and DeepSeek-Chat.
In the first round of self-repair, the models use a prompt that consists of the original prompt, the models' own generated specifications, an instruction to fix errors, and the error messages from the kernel verifier.
If the first round still produces incorrect specifications, in the second round of self-repair, the models use a prompt that consists of the prompt for the first round, the generated specifications for the first round, an instruction to fix errors, and the error messages from the kernel verifier for the first round.

\autotabref{table:self-repair} presents the results.
In total, GPT-4o repairs 14.67\% of erroneous specifications for the first round and 1.56\% for the second round, while DeepSeek-Chat repairs 11.45\% for the first round and 2.59\% for the second round.
In our experiments, self-repair
consistently improves the performance of LLMs in generating specifications for OS kernel verification. However, the repair success rate declines with additional rounds of repair.
Appendix~C in the extended version
presents a detailed case study on two representative error cases to investigate the root causes of these errors.

\subsection{Impact of Number of Demonstrations}

Recent studies have demonstrated that in-context learning (ICL) significantly enhances the ability of LLMs to acquire new tasks from a limited set of examples \cite{brown2020language,dong2022survey}. 
In the realm of OS verification, which is inherently complex, the provision of examples illustrating the generation of specifications from functional descriptions and code implementations exerts a substantial influence on performance outcomes.

We conduct experiments with five LLMs in various ICL settings, specifically 0-shot, 1-shot, 3-shot, and 5-shot learning.
We omit experiments with the other LLMs due to their prohibitive cost and time-intensive nature.

The 0-shot setting results in complete task failure, with a success rate of 0\% across all models, indicating the importance of demonstrations in OS verification contexts.

\autofigref{fig:few_shot_num} presents the results for the 1-, 3-, and 5-shot settings.
As the number of demonstrations increases, the performance across all five models improves by a large margin.
As anticipated, the \textit{Pass@1} performance of LLMs improves with the provision of additional demonstrations in our experiments. Notably, DeepSeek-Chat appears to derive greater benefits from increased demonstrations. While the reasoning model o3-mini surpasses the non-reasoning model DeepSeek-Chat in the 1-shot context, it underperforms DeepSeek-Chat in the 3- and 5-shot scenarios.
We hypothesize that advanced reasoning models used in our experiments generate extensive chains of reasoning traces, which may challenge the long-context learning capabilities in OS verification scenarios.

\subsection{Impact of Number of Injected Bugs}

We further investigate the impact of varying numbers of injected bugs on the performance of specification synthesis. \autofigref{fig:bug_num} presents the results. Our observations yield several insights:
1) In general, the \textit{Pass@1} performance of LLMs declines as the number of vulnerabilities increases.
We attribute this decline to the presence of more vulnerabilities in the kernel implementation, which complicates the models' ability to accurately comprehend the functional descriptions.
2) Advanced reasoning models underperform compared to traditional instruction-following models. For instance, GPT-4o consistently outperforms o1 and o3-mini across all levels of vulnerability. Similarly, DeepSeek-R1 is less effective than DeepSeek-Chat. These findings align with the results presented in \autotabref{table:main-results}.
These results indicate that reasoning models may encounter greater challenges due to the long-context limitations inherent in OS verification scenarios.


\section{Conclusion}

We introduce \Tool{}, a benchmark for evaluating the performance of LLMs in generating specifications for verifying OS kernels.
By formulating the specification generation task as a program synthesis problem, the benchmark challenges LLMs to navigate complex syntax and semantics within long-context tasks. 
Our comprehensive evaluation of 12 state-of-the-art LLMs reveals limitations in their current ability to handle these tasks effectively, with notable disparities in performance across models. 
These findings underscore the need for further advancements in LLM technology to enhance their understanding and generation capabilities in complex domains. 
\Tool{} not only highlights existing gaps but also serves as a valuable tool for guiding future research aimed at improving verification processes in operating system development.

\section*{Limitations}
\label{limitations}

\Tool{} is designed around Hyperkernel, which might not capture the full diversity of OS kernel architectures, potentially causing LLMs to overfit and limiting the generalizability of results to other systems.
The complexity of tasks, each consisting of approximately 20k to 30k tokens, poses significant challenges for LLMs in context management, which might overshadow other capabilities like logical reasoning.
The confined scope of syntax and semantics within the benchmark might not fully reflect the dynamic nature of real-world OS development environments. Current evaluation metrics might not capture qualitative aspects such as readability and adaptability. 
These limitations can guide future efforts to enhance benchmarks for evaluating LLMs in complex, real-world programming and verification tasks.

\section*{Acknowledgments}

We thank the anonymous reviewers, Sizhe Zhong, Huiri
Tan, and Jipeng Zhang for their insightful comments.
This work was supported in part by the Hong Kong Research Grants Council (Project No. 26216025) and Alibaba Group (through the Alibaba Innovative Research Program).
The views expressed in this work are those of the authors and do not necessarily reflect the views of the funding agencies.

\bibliography{aaai2026}

\clearpage
\appendix

\section{Synthesis Challenges}
\label{appendix:synthesis-challenges}

\begin{figure}[t]
    \centering
\includegraphics[width=0.6\linewidth]{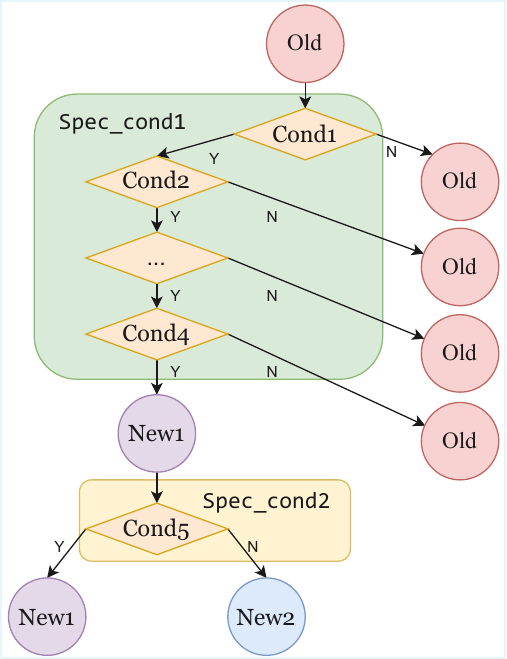}
    \caption{Divergent kernel states in the state-machine specification (\autofigref{fig:sys_close-spec}) of the system call \texttt{sys\_close} (\autofigref{fig:sys_close_impl}).
Nodes labeled \texttt{Old} represent the kernel state prior to the system call's execution, while nodes labeled \texttt{New1} or \texttt{New2} depict the potential kernel states following its successful completion. Rhombus-shaped nodes signify the conditions defined in the specification.
    }
    \label{fig:diverged-kernel-state}
\end{figure}

\noindent\textbf{Synthesize the correct specification with the corresponding semantics.}
\label{challenge:semantics-mapping}
In our formulation, the task of LLMs is to synthesize state-machine specifications within the semantics of the programming model, guided by the high-level functional description and code implementation of the system call.
To synthesize the correct specification from an informal functional description in natural language, the process must ensure accurate field access, appropriate constant selection, and precise condition selection. Specifically, to achieve deterministic specification synthesis, the synthesis process should operate within a symbolic triple relation, denoted as \textit{$\langle$ Desc, Impl, Model $\rangle$}.
This rule entails that to synthesize any given statement in the specification, the LLMs should first identify the relevant functional description (Desc), then locate the corresponding concrete kernel implementation (Impl).
Finally, they must interpret the intended semantics and synthesize the correct specification by referencing the appropriate model (Model) as defined within the programming model.

\begin{figure}[t]
    \centering
\begin{lstlisting}[language=python]
def sys_close(old, pid, fd):
    spec_cond1 = z3.And(
        z3.And(pid > 0, pid < dt.NPROC), # cond1.
        z3.And(fd >= 0, fd < dt.NOFILE), # cond2.
        z3.Or(
            pid == old.current,
            old.procs[pid].state == dt.proc_state.PROC_ZOMBIE), # cond3.
        z3.And(z3.UGT(old.procs[pid].ofile(fd), 0),
            z3.ULT(old.procs[pid].ofile(fd), dt.NFILE)) # cond4.
    )
    new1 = old.copy()
    fn = new1.procs[pid].ofile(fd)
    new1.procs[pid].ofile[fd] = z3.BitVecVal(0, dt.fn_t)
    new1.procs[pid].nr_fds[fd] -= 1
    new1.files[fn].refcnt[(pid, fd)]-= 1
    spec_cond2 = z3.And(new1.files[fn].refcnt() == 0) # cond5.
    new2 = new1.copy()
    new2.files[fn].type = dt.file_type.FD_NONE
    new2.files[fn].value = z3.BitVecVal(0, dt.uint64_t)
    new2.files[fn].offset = z3.BitVecVal(0, dt.off_t)
    new2.files[fn].omode = z3.BitVecVal(0, dt.uint64_t)
    new3 = util.If(spec_cond2,new2,new1)
    return spec_cond1, util.If(spec_cond1, new3, old)
\end{lstlisting}
    \caption{The state-machine specification of the system call \texttt{sys\_close} (\autofigref{fig:sys_close_impl}). In this context, \texttt{old} represents the initial kernel state, while \texttt{new1} to \texttt{new3} denote the subsequent kernel states that the kernel will transition to upon successful completion of the system call.
}
    \label{fig:sys_close-spec}
\end{figure}

For example, \autofigref{fig:sys_close_impl} shows part of the implementation of the system call \texttt{sys\_close}.
As the functional description indicates, ``This involves updating the process's file descriptor table to mark the descriptor as unused and decrementing the count of open file descriptors for the process. Additionally, the system updates the file's reference count to reflect the decrease in file references.''
The LLMs then locate the relevant kernel code implementation on lines 12-14 of \autofigref{fig:sys_close_impl}, as shown in \autofigref{lst:impl}.

The LLMs then synthesize the state-machine specification based on the programming model in \autofigref{fig:programming-model} and the abstract syntax in \autofigref{fig:spec:domain} (in the main paper).
The synthesized state-machine specification should correspond to lines 13-15 in \autofigref{fig:sys_close-spec}, as shown in \autofigref{lst:spec}.

Here, the LLMs interpret the semantics of terms such as ``the process's file descriptor table'' as \texttt{new1.procs[pid].ofile}, ``the count of open file descriptors for the process'' as \texttt{new1.procs[pid].nr\_fds[fd]}, and ``the file's reference count'' as \texttt{new1.files[fn].refcnt[(pid, fd)]} accordingly. In this context, \texttt{new1} denotes the kernel state, while \texttt{new1.procs[pid]} specifies the process, among other mappings.

\noindent\textbf{Divergent kernel states synthesis.}
\label{challenge:kernel-states-synthesis}
To synthesize an accurate state-machine specification, it is essential to determine the kernel state to which the system transitions under specific conditions according to the abstract syntax in \autofigref{fig:spec:domain} (in the main paper).

This process requires LLMs to perform advanced reasoning, as many system calls involve cascading and interdependent conditions.
\autofigref{fig:diverged-kernel-state} illustrates the divergent kernel states resulting from the system call \texttt{sys\_close}. Specifically, it indicates that the kernel transitions to state \texttt{new1} when both condition \texttt{spec\_cond1} and condition \texttt{spec\_cond2} are satisfied. If condition \texttt{spec\_cond1} is satisfied but condition \texttt{spec\_cond2} is not, the kernel transitions to state \texttt{new2}. Finally, if neither condition \texttt{spec\_cond1} nor condition \texttt{spec\_cond2} is satisfied, the kernel remains in state \texttt{old}.

\begin{figure}[t]
    \centering
\begin{lstlisting}[language=python] 
class KernelState(BaseStruct):
    procs = Proc()
    pages = Page()
    files = File()
    pcipages = PCIPage()
    pci = PCI()
    ...
class Proc(Struct):
    ...
    ofile = Map(pid_t, fd_t, fn_t)
    nr_fds = Refcnt(pid_t, fd_t, size_t)
class File(Struct):
    ...
    refcnt = Refcnt(fn_t, (pid_t, fd_t), size_t)
class PCIPage(Struct):
    owner = Map(pn_t, devid_t)
    valid = Map(pn_t, bool_t)
class PCI(Struct):
    ...
    owner = Map(devid_t, pid_t)
\end{lstlisting}
\caption{Part of the programming model.}
\label{fig:programming-model}
\end{figure}

\begin{figure}[t]
\centering
\begin{lstlisting}[language=python]
new1.procs[pid].ofile[fd] = z3.BitVecVal(0, dt.fn_t)
new1.procs[pid].nr_fds[fd] -= 1
new1.files[fn].refcnt[(pid, fd)] -= 1
\end{lstlisting}
\caption{Relevant state-machine specification.}
\label{lst:spec}
\end{figure}

\begin{figure}[t]
    \centering
\begin{lstlisting}[language=c]
int sys_close(pid_t pid, int fd) {
    if (!is_pid_valid(pid)) // Cond1.
        return -ESRCH;
    if (!is_fd_valid(fd))   // Cond2.
        return -EBADF;
    ...
    clear_fd(pid, fd);
    ...
}
static inline void clear_fd(pid_t pid, int fd) {
    ...
    file = get_file(get_fd(pid, fd));
    proc->ofile[fd] = 0;
    --proc->nr_fds;
    if (--file->refcnt == 0) { // Cond5.
        ...
    }
}
\end{lstlisting}
    \caption{Part of the code implementation of the system call \texttt{sys\_close}.}
    \label{fig:sys_close_impl}
\end{figure}

\begin{figure}[t]
    \centering
\begin{lstlisting}[language=c]
proc->ofile[fd] = 0;
--proc->nr_fds;
--file->refcnt;
\end{lstlisting}
\caption{Relevant potentially buggy code implementation according to the functional description.}
\label{lst:impl}
\end{figure}

\section{Examples of Five Bug Types}\label{sec:bug_def}

We selected five types of bugs from the xv6 kernel \cite{cox2011xv6} to inject into the kernel implementation, as illustrated below.
These types of vulnerabilities represent common and critical types of vulnerabilities frequently observed in the OS kernel domain, for example, in the Linux kernel \cite{linux_cve_announce}.

\begin{figure*}[t]
    \centering
    \includegraphics[width=\linewidth]{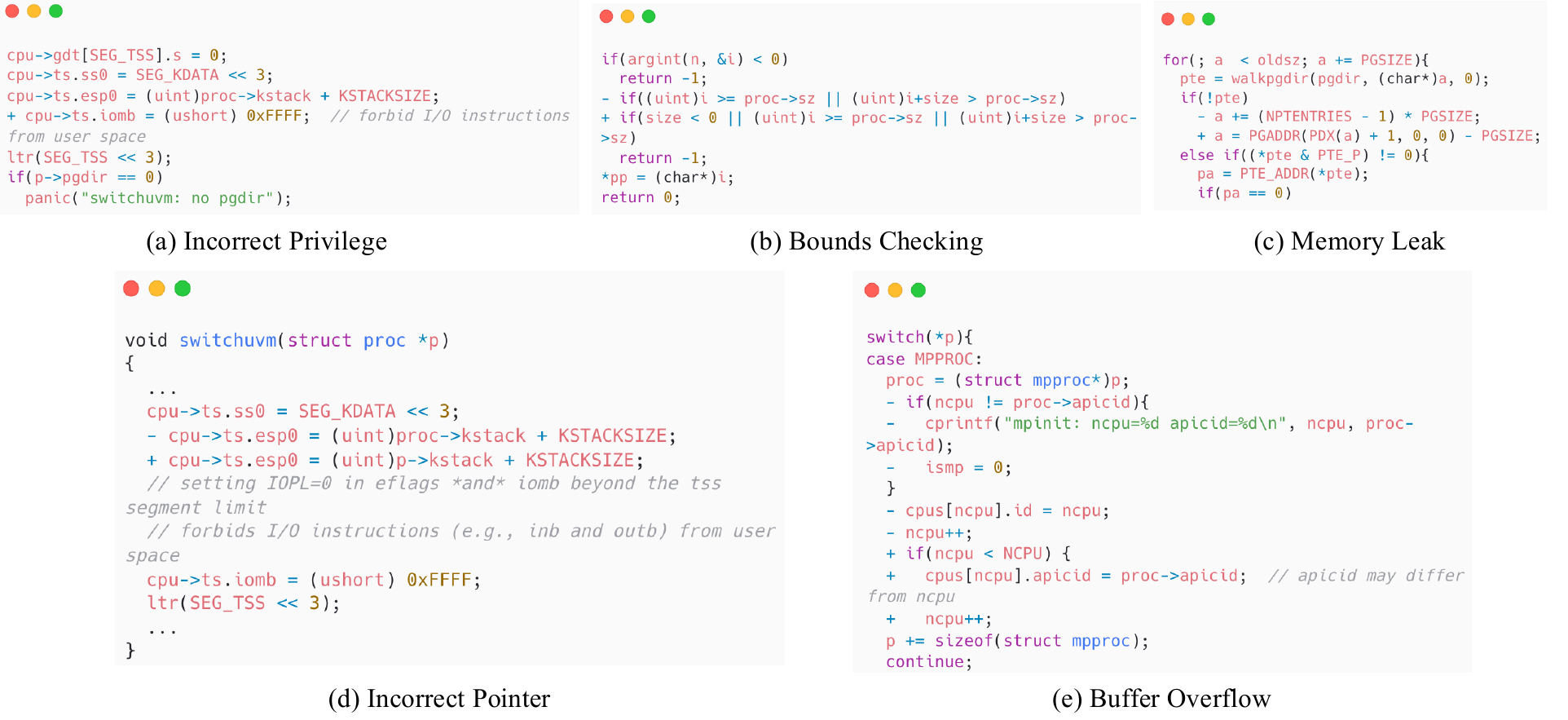}
    \caption{Illustrative examples of five bug types extracted from the xv6 kernel \cite{cox2011xv6}.}
    \label{fig:bug-types}
\end{figure*}

\noindent\textbf{Incorrect privilege.}
As shown in \autofigref{fig:bug-types} (a), the incorrect privilege bug occurs because the kernel fails to set the \texttt{iomb} field in the Task State Segment (TSS), leaving it at its default value (0), which allows user-space processes to execute I/O instructions directly. This violates privilege separation, enabling malicious processes to bypass kernel control, access hardware, corrupt device state, or destabilize the system.

\noindent\textbf{Bounds checking.}
As shown in \autofigref{fig:bug-types} (b), the bounds-checking bug occurs because the second condition fails to check whether \texttt{size} is negative. Without this check, a negative size can cause an integer underflow in \texttt{(uint)i + size}, bypassing the bounds check and allowing invalid memory access. This results in potential memory violations, including accessing or modifying out-of-bounds memory, leading to undefined behavior.

\noindent\textbf{Memory leak.}
As illustrated in \autofigref{fig:bug-types} (c), the memory leak bug arises from a faulty statement that causes the search to resume at the first entry of the next page table, instead of the current entry, when a zero page directory entry (PDE) is encountered while searching for present page table entries (PTEs) to free.
This leads to leaked memory, which remains allocated but unusable, potentially exhausting system resources over time.

\noindent\textbf{Incorrect pointer.}
As shown in \autofigref{fig:bug-types} (d), the incorrect pointer bug occurs because the switchuvm() function is intended to switch the Task State Segment (TSS) and page table to the process p that is passed as an argument. However, instead of using p to access the kstack field, the function erroneously references the global proc. This misuse of the pointer results in an incorrect kernel state transition.

\noindent\textbf{Buffer overflow.}
As shown in \autofigref{fig:bug-types} (e), the buffer overflow bug occurs because the kernel incorrectly assumes that \texttt{cpu->id} values (APIC IDs) are consecutive and start from zero, using them directly as indices into the cpus array.
If APIC IDs are sparse or non-consecutive, this results in an out-of-bounds memory access, potentially corrupting kernel memory, causing system instability, or introducing security vulnerabilities.

\section{Case Study}
\label{sec:case_study}

We next conduct a case study to further investigate the root causes of these errors.

\noindent\textbf{Syntax error.}
First, we present an example of a syntax error. This error occurs during the synthesis of the state-machine specification for the system call \texttt{sys\_map\_pcipage}.
The functional description of this system call is as follows: it verifies that the current process owns the specified PCI page.
The corresponding code snippet is presented in \autofigref{lst:impl-pcipage}.

\begin{figure}[t]
    \centering
\begin{lstlisting}[language=c]
int sys_map_pcipage(pn_t pt, size_t index, pn_t pcipn, pte_t perm) {
    if (!is_pcipn_pid(pcipn, pid))
        return -EACCES;
}
static int is_pcipn_pid(pn_t pcipn, pid_t pid) {
    ...
    return pci_table[pcipages[pcipn].devid] == pid;
}
\end{lstlisting}
\caption{Part of the code implementation of the system call \texttt{sys\_map\_pcipage}.}
\label{lst:impl-pcipage}
\end{figure}

The LLMs then synthesize the specification, as presented in \autofigref{lst:spec-pcipage}.

\begin{figure}[t]
    \centering
\begin{lstlisting}[language=python]
...
old.pcipages[pcipn].devid == old.current
...
\end{lstlisting}
\caption{Part of the incorrect state-machine specification of system call \texttt{sys\_map\_pcipage}.}
\label{lst:spec-pcipage}
\end{figure}

However, based on the provided programming model \autofigref{fig:programming-model}, \texttt{old.pcipages[pcipn]} is intended to retrieve a modeled instance of a \texttt{PCIPage}, which does not include a modeled \texttt{devid} attribute.
This error results in an attempt to access a non-existent class field in Python, leading to a syntax error.

This failure may be attributed to the long context, which might cause the LLMs to lose track of the provided programming model and instead synthesize a specification resembling the concrete code implementation of the system call. However, it is crucial to emphasize that the kernel's abstract modeling differs from its code implementation.

\noindent\textbf{Semantic error.}
Next, we present another example to demonstrate the occurrence of semantic errors. This example involves a mistake in the synthesis of the state-machine specification for the system call \texttt{sys\_map\_proc}. The functional description of this system call is as follows: it verifies the specified permissions and rejects the operation if the permissions include write access. However, the code implementation contains an injected bug, as shown in \autofigref{lst:impl-map-proc}.
\begin{figure}[t]
    \centering
\begin{lstlisting}[language=c]
- if (pte_writable(perm)) [correct] 
+ if (!pte_writable(perm))[bug injected]
    return -EACCES;
static inline bool pte_writable(uintptr_t x)
{
    return x & PTE_W;
}
\end{lstlisting}
\caption{Part of the code implementation of the system call \texttt{sys\_map\_proc}.}
\label{lst:impl-map-proc}
\end{figure}

\begin{figure}[t]
\begin{lstlisting}[language=python]
...
perm & dt.PTE_W != 0,
...
\end{lstlisting}
\caption{Part of the correct state-machine specification of system call \texttt{sys\_map\_proc}.}
\label{lst:proc-spec-c}
\end{figure}

\begin{figure}[t]
    \centering
\begin{lstlisting}[language=python]
...
perm & dt.PTE_W == 0,
...
\end{lstlisting}
\caption{Part of the incorrect state-machine specification of system call \texttt{sys\_map\_proc}.}
\label{lst:proc-spec-i}
\end{figure}

In this case, the LLMs are expected to identify the injected vulnerability by adhering to the functional description and synthesizing the correct specification, as shown in \autofigref{lst:proc-spec-c}.

The LLMs, however, fail to accurately interpret the functional description, resulting in the generation of an incorrect specification, as shown in \autofigref{lst:proc-spec-i}.

This incorrect condition causes the kernel state to transition into an inconsistent state that deviates from the correct operating system implementation, ultimately resulting in a semantic error.

\section{Task Prompt Example}
\label{example:task-prompt}

Here we provide a full prompt example for a task below:

\begin{tcolorbox}[title=System, systembox]
You will be provided with four sections displayed below, including verification assumption, programming model, a few examples, and a task question.
The operating system kernel verification will be performed under the verification assumption.
...
Then, you will be given with a user question that includes the functional description and a potentially buggy code implementation of a system call. Your task is to synthesize the corresponding state-machine specification that verifies the functional correctness of the given system call.
\end{tcolorbox}

\begin{tcolorbox}[title=Verification Assumption, assumptionbox]
The kernel provides the abstraction of a process using Intel VT-x and AMD-V virtualization support. It runs as a host and user processes run as guests (in ring 0). Trap handlers are implemented as VM-exit handlers, in response to hypercalls (to implement system calls), preemption timer expiration, exceptions, and interrupts. The approach firstly allows the kernel and user space to have separate page tables; the kernel simply uses an identity mapping for its own address space, sidestepping reasoning about virtual-to-physical mapping for kernel code. Secondly, the use of virtualization safely exposes the interrupt descriptor table (IDT) to user processes. This allows the CPU to deliver exceptions (e.g. general protection or page fault) directly to user space, removing the kernel from most exception-handling paths. The kernel has some limitations. It does not support threads, copy-on-write fork, shared pages, or Unix permissions.

...
\end{tcolorbox}

\begin{tcolorbox}[title=Programming Model, programmingbox]
Constants available:
\begin{pycode}
intremap_state = {
    IR_FREE: z3.BitVecVal(0, 64),
    IR_FORCE_WIDTH: z3.BitVecVal(
          18446744073709551615, 64),
    IR_ACTIVE: z3.BitVecVal(1, 64)}
...
PAGE_SIZE = 4096
PCI_START = 0xa0000000
PCI_END = 0x100000000
...
intremap_state_t = z3.BitVecSort(64)
file_type_t = z3.BitVecSort(64)
proc_state_t = z3.BitVecSort(64)
page_type_t = z3.BitVecSort(64)
...

\end{pycode}
Class models:
\begin{pycode}
class KernelState(BaseStruct):
    procs = Proc()
    dmapages = DMAPage()
    files = File()
    pci = PCI()
    ...
class File(Struct):
    type = Map(fn_t, file_type_t)
    offset = Map(fn_t, size_t)
    ...
class Proc(Struct):
    state = Map(pid_t, proc_state_t)
    killed = Map(pid_t, bool_t)
    ...
class Page(Struct):
    data = Map(pn_t, uint64_t, uint64_t)
    owner = Map(pn_t, pid_t)
    type = Map(pn_t, page_type_t)
    ...
class DMAPage(Struct):
    owner = Map(pn_t, pid_t)
    type = Map(pn_t, page_type_t)
    ...
class PCIPage(Struct):
    owner = Map(pn_t, devid_t)
    valid = Map(pn_t, bool_t)
    ...
class PCI(Struct):
    owner = Map(devid_t, pid_t)
    page_table_root = Map(devid_t, pn_t)
    ...
...
\end{pycode}
External functions: Other than the model above, z3 datatypes and methods are included in the model, such as z3.ULT(), z3.UGT(), z3.ULE(), z3.UGE(), etc. ... \\
...
\end{tcolorbox}

\begin{tcolorbox}[title=Examples, examplesbox]
Example 1: Given a system call sys\_alloc\_port. \\
\textbf{Functional Description}: The sys\_alloc\_port system call is designed to allocate an I/O port to a process, allowing it to perform input/output operations on that port. This functionality is crucial in systems where processes need direct access to hardware resources, such as peripheral devices, through specific I/O ports.
...
The system call begins by checking if the requested port is already in use. If the port is occupied, the operation is terminated with an error, indicating that the port is unavailable for allocation. This check prevents conflicts and ensures that each port is uniquely assigned to a single process at any given time. \\
... \\
\textbf{Code Implementation}:   \\
Its corresponding code implementation that may contain bugs is as follows: 
\begin{pycode}
int sys_alloc_port(uint16_t port)
{
    struct proc *proc;
    if (io_table[port])
        return -EBUSY;
    proc = get_proc(current);
    if (!proc->use_io_bitmap)
        return -EACCES;
    io_table[port] = current;
    if (port < 0x8000)
        bit_clear(port,
                get_page(proc->io_bitmap_a));
    else
        bit_clear(port - 0x8000,
                get_page(proc->io_bitmap_b));
    ...
    return 0;
}
...
\end{pycode}

\textbf{Specification}: Based on the detailed functional description and the potentially buggy code implementation of the system call sys\_alloc\_port provided above, the state-machine specification of the system call is deduced as follows: 
\begin{pycode}
def sys_alloc_port(old, port):
    cond = z3.And(
        old.io[port].owner == 0,
        old.procs[old.current].use_io_bitmap,
    )

    new = old.copy()

    new.io[port].owner = old.current
    new.procs[old.current].nr_ports[port] += 1

    page = util.If(z3.ULT(port, 0x8000),
            new.procs[new.current].io_bitmap_a,
            new.procs[new.current].io_bitmap_b)

    ...
    return cond,util.If(cond, new, old)
\end{pycode}
Example n: ...
\end{tcolorbox}

\begin{tcolorbox}[title=Task Question, questionbox]
Now, given the system call sys\_close.  \\
\textbf{Functional Description}: 
The sys\_close system call is responsible for closing a file descriptor associated with a specific process, thereby releasing the resources tied to that descriptor. \\
...
\\
The system call begins by validating the process identifier to ensure it corresponds to a valid process.    \\
... \\
\textbf{Code Implementation}: Its corresponding code implementation that may contain bugs is as follows: 
\begin{pycode}
int sys_close(pid_t pid, int fd)
{
    if (!is_pid_valid(pid))
        return -ESRCH;
    if (!is_fd_valid(fd))
        return -EBADF;
    if (pid != current &&
      !is_proc_state(pid, PROC_ZOMBIE))
        return -EACCES;
    if (get_fd(pid, fd) == 0)
    ...
    return 0;
}
...
\end{pycode}
\textbf{Specification}: Based on the detailed functional description and the potentially buggy code implementation of the system call sys\_close provided above, the state-machine specification of the system call is deduced as follows: 
\end{tcolorbox}

Given the task prompt provided above, the LLMs are expected to synthesize the state-machine specification for the system call \texttt{sys\_close} as follows:

\begin{tcolorbox}[title=LLM Output, answerbox]
\begin{pycode}
def sys_close(old, pid, fd):
 spec_cond1 = z3.And(
   z3.And(pid > 0, pid < dt.NPROC),
   z3.And(fd >= 0, fd < dt.NOFILE),
   z3.Or(pid == old.current, old.procs[pid].state
                  == dt.proc_state.PROC_ZOMBIE),
   z3.And(z3.UGT(old.procs[pid].ofile(fd), 0),
       z3.ULT(old.procs[pid].ofile(fd),dt.NFILE))
 )
 new1 = old.copy()
 fn = new1.procs[pid].ofile(fd)
 new1.procs[pid].ofile[fd]=z3.BitVecVal(0,dt.fn_t)
 new1.procs[pid].nr_fds[fd] -= 1
 new1.files[fn].refcnt[(pid, fd)] -= 1
 spec_cond2 = z3.And(new1.files[fn].refcnt() == 0)
 new2 = new1.copy()
 new2.files[fn].type = dt.file_type.FD_NONE
 new2.files[fn].value=z3.BitVecVal(0,dt.uint64_t)
 new2.files[fn].offset = z3.BitVecVal(0, dt.off_t)
 new2.files[fn].omode=z3.BitVecVal(0,dt.uint64_t)
 new3 = util.If(spec_cond2, new2, new1)
 return spec_cond1, util.If(spec_cond1, new3, old)
\end{pycode}
\end{tcolorbox}

\section{Benchmark Analysis}
\label{appendix:benchmark-analysis}

\begin{table}[t]
    \centering
    \caption{The number of tokens for benchmark tasks. The \textbf{Prompt} column specifies the type of prompt utilized for each task, including state-machine specification generation prompts with 1-shot, 3-shot, and 5-shot examples, as well as specification repair prompts. The columns \textbf{Min}, \textbf{Max}, \textbf{Median}, and \textbf{Mean} represent the respective statistical metrics: minimum, maximum, median, and mean values.
    }
    \resizebox{\linewidth}{!}{
    \begin{tabular}{l|cccc}
        \toprule
        Prompt & \makecell{Min} & \makecell{Max} & \makecell[c]{Median} &  \makecell[c]{Mean} \\ 
        \midrule
        1-shot prompt & 12,604 & 14,619 & 13,740 & 13,619 \\ 
        3-shots prompt & 19,282 & 21,297 & 20,418 & 20,297 \\ 
        5-shots prompt & 23,259 & 25,274 & 24,395 & 24,274 \\ 
        5-shots repair prompt & 23,833 & 34,228 & 26,796 & 27,210 \\
        \bottomrule
    \end{tabular}
    }
\label{table:token-stats}
\hfill
    \centering
    \caption{
    The number of tasks associated with each type of vulnerability. The column \textbf{Bug Type} specifies the category of vulnerabilities, while the column \textbf{Task} indicates tasks in which the code implementation of the system call contains the corresponding type of vulnerability.
    }
    \begin{tabular}{l|c}
        \toprule
        Bug type & \makecell{Tasks} \\ 
        \midrule
        Incorrect pointer & 71 \\ 
        Incorrect privilege & 112 \\ 
        Buffer overflow & 54 \\ 
        Memory leak & 74 \\ 
        Bounds checking & 108 \\ 
        \bottomrule
    \end{tabular}
\label{table:bug-type-task-stats}
\end{table}

\begin{figure}[!t]
    \centering
    \begin{subfigure}[b]{0.48\textwidth}
      \centering
      \includegraphics[width=\linewidth]{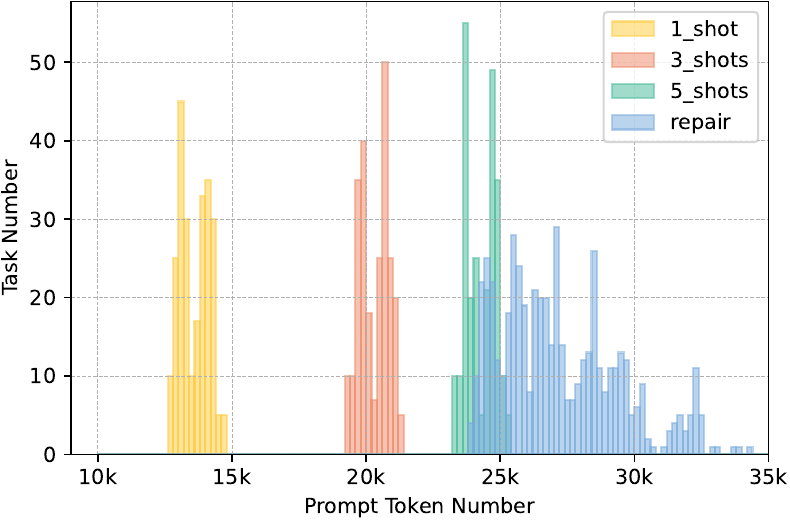}
      \caption{Distribution of task prompt token numbers. \textbf{1\_shot}, \textbf{3\_shots}, and \textbf{5\_shots} refer to prompts containing 1-shot, 3-shot, and 5-shot examples, respectively, for specification generation tasks, while \textbf{repair} denotes prompts with 5-shot examples used for specification repair tasks.}
      \label{fig:token-distribution}
    \end{subfigure}
    \hfill
    \begin{subfigure}[b]{0.48\textwidth}
      \centering
      \includegraphics[width=0.8\linewidth, clip, trim={15pt 15pt 15pt 15pt}]{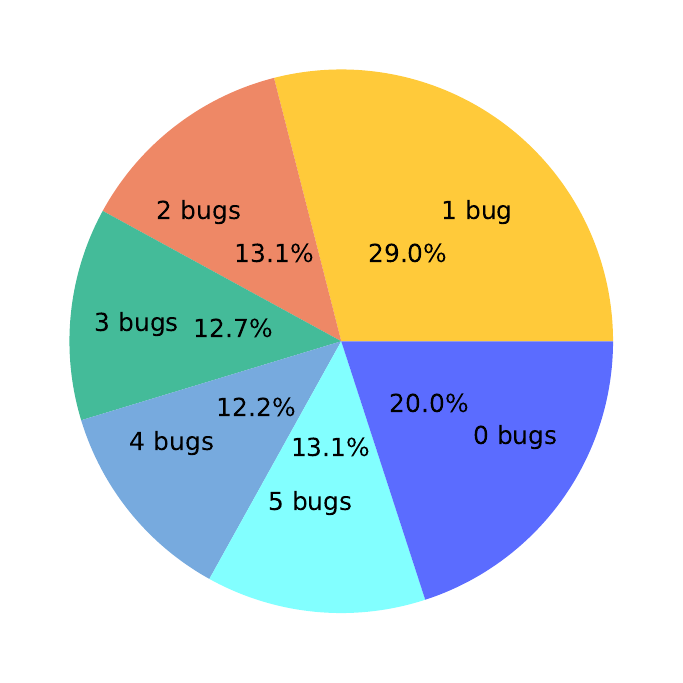}
      \caption{Percentage of tasks containing code implementations of system calls with varying numbers of injected vulnerabilities. The labels from \textbf{0 bugs} to \textbf{5 bugs} denote tasks with code implementations of system calls that include the corresponding number of injected vulnerabilities.}
      \label{fig:bug-num-pie}
    \end{subfigure}
    \caption{Statistics.}
  \end{figure}

We conducted a comprehensive analysis of the benchmark under study. Initially, we performed basic statistical analyses of the number of tokens within the prompts for both the specification generation tasks and the specification repair tasks, calculated using \cite{tiktoken}, as shown in \autotabref{table:token-stats}. To make the statistics more intuitive, we further analyzed the distribution of token counts across task prompts, which is illustrated in \autofigref{fig:token-distribution}.
The histogram and table reveal that the token counts of task prompts range from 10k to 35k tokens.
Under the standard 5-shot example setting, the majority of task prompts are observed to fall within the range of 20k to 30k tokens.

\autotabref{table:bug-type-task-stats} summarizes the number of tasks corresponding to each type of vulnerability. It can be observed that tasks involving code implementations of system calls with the vulnerability \textbf{Incorrect Pointer} are the most prevalent, with a total of 112 instances, whereas tasks associated with \textbf{Buffer Overflow} are the least prevalent, with only 54 instances.

\autofigref{fig:bug-num-pie} illustrates the distribution of tasks based on the percentage of code implementations of system calls containing varying numbers of injected vulnerabilities. The analysis reveals that implementations with a single injected bug constitute the largest proportion, accounting for 29.0\%, while those containing four injected bugs are the least common, representing only 12.2\%.

\end{document}